\PassOptionsToPackage{colorlinks,bookmarksopen,bookmarksnumbered,citecolor=red,urlcolor=red}{hyperref}
\documentclass[a4paper,fleqn,authoryear]{cas-dc}
\usepackage[numbers]{natbib}
\usepackage{color}
\usepackage{colortbl}
\usepackage{soul}  
\usepackage{xcolor}  
\usepackage{seqsplit} 
\usepackage{float}
\usepackage{amssymb}
\usepackage{hyperref}      
\usepackage{cleveref}      
\usepackage{pifont}
\usepackage{pifont} 
\usepackage{tikz}
\usepackage{float}     
\usepackage{placeins}  
\usepackage{caption} 
\usepackage{natbib}
\usepackage{xcolor}
\usetikzlibrary{arrows.meta, positioning, shapes.geometric, shapes.misc, fit, calc}

\tikzset{
  >={Latex[length=2mm]},
  box/.style = {rectangle, rounded corners=2mm, draw, thick, align=center, minimum height=10mm, minimum width=22mm, fill=white},
  smallbox/.style = {rectangle, rounded corners=1mm, draw, align=center, minimum height=8mm, minimum width=18mm, fill=white},
  decision/.style = {diamond, draw, thick, aspect=2, align=center, inner sep=1.5pt, minimum size=10mm},
  note/.style = {draw, dashed, rounded corners=2mm, inner sep=2mm},
  arrow/.style = {->, thick},
  lab/.style = {font=\footnotesize, inner sep=0.5pt},
}

\definecolor{uclablue}{rgb}{0.15, 0.45, 0.68}
\hypersetup{
    citecolor=uclablue,
    colorlinks=true,
}
\definecolor{my_green}{RGB}{51,102,0}
\definecolor{my_red}{RGB}{204, 0, 0}

\def\tsc#1{\csdef{#1}{\textsc{\lowercase{#1}}\xspace}}
\tsc{WGM}
\tsc{QE}
\tsc{EP}
\tsc{PMS}
\tsc{BEC}
\tsc{DE}

\begin{document}
\let\WriteBookmarks\relax
\def\floatpagepagefraction{1}
\def\textpagefraction{.001}
\shorttitle{ElectriQ: Benchmarking LLMs for Power Marketing}
\shortauthors{Jinzhi Wang et~al.}

\title [mode = title]{ElectriQ: LLM Dialogue Benchmark for Electric Power Marketing in Renewable-Integrated Power Systems}                      
\tnotemark[1,2]

\tnotetext[1]{This work was supported by the National Natural Science Foundation of China under Grant No. 61872288.
}
\tnotetext[2]{The data and code is publicly available at: \url{https://github.com/Jinzhi-Wang/ElectriQ}.
}

\author[1, 2]{Jinzhi Wang}[type=author,
                   auid=001,
                   bioid=1,
                   style=chinese,
                   orcid=0009-0008-8964-7164]
\ead{wangjz5515@stu.xjtu.edu.cn}
\credit{Conceptualization, Methodology, Formal analysis, Software, Data curation, Validation, Writing – original draft, Visualization, Writing – review \& editing}
\author[1]{Qingke Peng}[type=author,
                   auid=003,
                   bioid=3,
                   style=chinese,
                   orcid=0000-0002-5448-8529]
\cormark[1]
\ead{qkpeng@mail.xjtu.edu.cn}
\credit{Conceptualization, Supervision, Methodology, Project administration, Writing – review \& editing, Funding acquisition}
\author[1]{Haozhou Li}[type=author,
                   auid=004,
                   bioid=4,
                   style=chinese]
\ead{lihaozhou1126@stu.xjtu.edu.cn}
\credit{Conceptualization, Supervision, Writing – review \& editing }
\author[1]{Zeyuan Zeng}[type=author,
                   auid=005,
                   bioid=5,
                   style=chinese]
\ead{zengzeyuan@stu.xjtu.edu.cn}
\credit{Data curation, Software}
\author[4]{Jiangbo Zhang}[type=author,
                   auid=008,
                   bioid=8,
                   style=chinese]
\ead{4522253009@stu.xjtu.edu.cn}
\credit{Data curation, Methodology, Writing – original draft}
\author[3]{Kaixuan Yang}[type=author,
                   auid=007,
                   bioid=7,
                   style=chinese]
\ead{yangkaixuan1@stu.xjtu.edu.cn}
\credit{Formal analysis, Validation}
\author[5]{Ningyong Wu}[type=author,
                   auid=0010,
                   bioid=10,
                   style=chinese]
\ead{ninglongwu@stu.xjtu.edu.cn}
\credit{Data curation, Visualization}
\author[2]{Qinfeng Song}[type=author,
                         auid=002,
                         bioid=2,
                         style=chinese,
                         orcid=0000-0002-0721-8972]
\ead{sqfbjeb2361@gmail.com}
\credit{Investigation, Writing – review \& editing}
\author[1]{Ruimeng Li}[type=author,
                   auid=006,
                   bioid=6,
                   style=chinese]
\ead{lrm1105@stu.xjtu.edu.cn}
\credit{Writing – original draft, Validation}
\author[1]{Biyi Zhou}[type=author,
                   auid=009,
                   bioid=9,
                   style=chinese]
\ead{zhoubiyi@stu.xjtu.edu.cn}
\credit{Formal analysis, Validation}
\affiliation[1]{organization={Systems Engineering Institute, Xi’an Jiaotong University},
                addressline={No.28, Xianning West Road},
                city={Xi’an},
                postcode={710049},
                country={China}}
\affiliation[2]{organization={State Grid Hefei Electric Power Supply Company},
                addressline={No.133 Susong Road, Baohe District},
                city={Hefei},
                postcode={230022},
                country={China}}
\affiliation[3]{organization={State Key Laboratory of Multiphase Flow in Power Engineering, School of Energy and Power Engineering, Xi’an Jiaotong University},
                addressline={No.28, Xianning West Road},
                city={Xi’an},
                postcode={710049},
                country={China}}
\affiliation[4]{organization={School of Electronic Science and Engineering, Xi'an Jiaotong University},
                addressline={No.28, Xianning West Road, Beilin District},
                city={Xi'an},
                postcode={710049},
                country={China}}

\affiliation[5]{organization={Organizational Management Department, School of Management, Xi’an Jiaotong University},
                addressline={No.28, Xianning West Road, Beilin District},
                city={Xi’an},
                postcode={710049},
                country={China}}

\cortext[cor1]{Corresponding author}
\begin{abstract}
As power systems decarbonise and digitalise, high penetrations of distributed energy resources and flexible tariffs make electric power marketing (EPM) a key interface between regulation, system operation and sustainable-energy deployment. Many utilities still rely on human agents and rule- or intent-based chatbots with fragmented knowledge bases that struggle with long, cross-scenario dialogues and fall short of requirements for compliant, verifiable and DR-ready interactions. Meanwhile, frontier large language models (LLMs) show strong conversational ability but are evaluated on generic benchmarks that underweight sector-specific terminology, regulatory reasoning and multi-turn process stability. To address this gap, we present ElectriQ, a large-scale benchmark and evaluation framework for LLMs in EPM. ElectriQ contains over 550k dialogues across six service domains and 24 sub-scenarios and defines a unified protocol that combines human ratings, automatic metrics and two compliance stress tests—Statutory Citation Correctness and Long-Dialogue Consistency. Building on ElectriQ, we propose SEEK-RAG, a retrieval-augmented method that injects policy and domain knowledge during finetuning and inference. Experiments on 13 LLMs show that domain-aligned 7B models with SEEK-RAG match or surpass much larger models while reducing computational cost, providing an auditable, regulation-aware basis for deploying LLM-based EPM assistants that support demand-side management, renewable integration and resilient grid operation.
\end{abstract}



\maketitle
\section{Introduction}

As the global energy transition towards carbon neutrality and deep digitalisation accelerates, power systems are shifting from supply-side dominance to coordinated “source–grid–load–storage” operation with much stronger user participation \cite{Zhu_Electronics_SGLS_2024}. High penetrations of distributed photovoltaics and storage, the rapid growth of electric-vehicle (EV) charging \cite{IEA_GEV_2025}, and the widespread adoption of time-of-use (TOU) tariffs \cite{Leon_SETA_V2H_TOU_2023} and demand-response (DR) programmes \cite{TANG2024892} are reshaping how modern power systems are planned and operated. In this context, electric power marketing (EPM) has become a critical socio-technical interface between regulation, system operation and sustainable-energy deployment \cite{Mutarraf_SETA_EVInfra_2022}. Through hotlines, web portals and mobile applications, EPM delivers high-frequency services such as tariff consultation, metering and billing, energy-efficiency guidance, DR enrolment and coaching, distributed energy resource (DER) interconnection support, and outage or restoration communication \cite{Bhattacharyya_SciRep_2023}. The quality and consistency of these interactions directly affect demand-side flexibility, DER hosting capacity, grid resilience and, ultimately, achievable reductions in carbon intensity \cite{Hisoglu_SETA_SolarEVCS_2025}.

\begin{figure}
    \centering
    \includegraphics[width=1\linewidth]{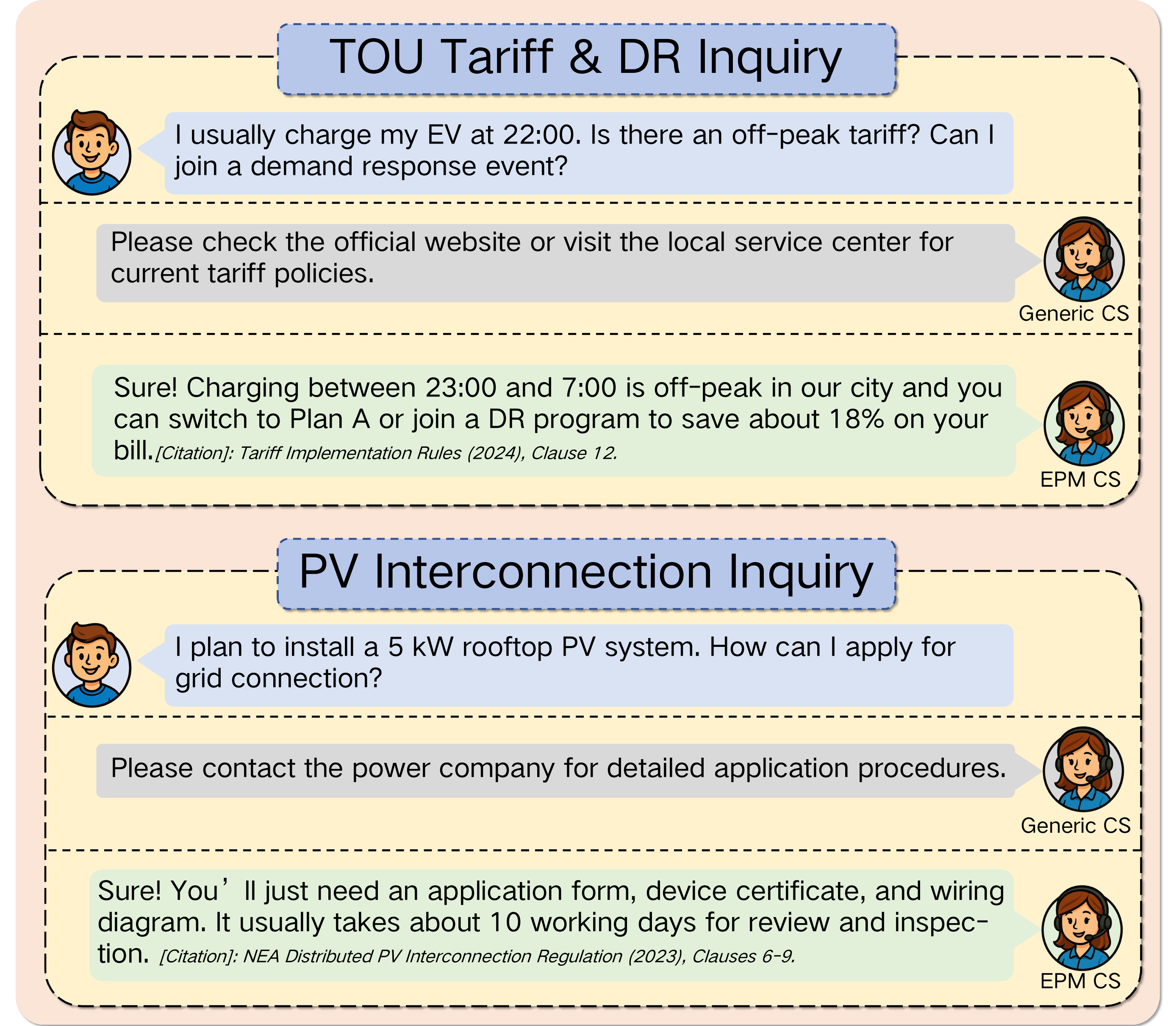}
    \caption{\textbf{General customer service vs.\ power-marketing dialogue. }
EPM dialogues must integrate pricing policies, consumption patterns, and regulatory constraints to produce actionable, compliant responses, motivating a dedicated benchmark.}
    \label{fig1}
\end{figure}

However, much of the demand-side management (DSM) \cite{PANDA20232228} and flexibility literature implicitly assumes that tariffs, contracts and interconnection procedures are perfectly understood and implemented at the customer interface, leaving an implementation gap between modelled benefits and real-world operation \cite{NEBEY20245422}. In practice, customer service in many utilities is still dominated by human agents and rule- or template-based IVR/FAQ/intent-recognition chatbots \cite{Følstad_Chatbots_Agenda_2021}, which rely on shallow semantic matching over fragmented, asynchronously updated knowledge bases \cite{Mo_CSurv_ConversationalSearch_2025,Mariani_JBR_CA_2023}. These systems struggle with cross-scenario, long-turn dialogues that mix TOU tariffs, DR incentives and DER interconnection rules, and consequently fall short of sustainable-energy requirements for technically accurate, auditable and DR-ready interactions \cite{Parrish_EnergyPolicy_DR_2020}: misinterpreted tariffs undermine TOU adherence, inconsistent DR explanations depress enrolment, and non-compliant interconnection guidance slows DER adoption and increases regulatory risk. As illustrated in Fig.~\ref{fig1}, effective EPM assistants must integrate pricing policies, consumption patterns and regulatory constraints to produce actionable, compliant responses, motivating the need for a dedicated benchmark.

Frontier large language models (LLMs) such as GPT-4 and Claude, together with strong open-source counterparts, support multi-turn dialogue and retrieval-augmented generation, and early prototypes in sustainable-energy contexts have explored applications in sustainability assessment \cite{Li_SustainLLM_SETA_2025}, personalised energy advice, grid-operation assistance \cite{Cheng_SciRep_2025}, dynamic tariff recommendation and DR optimisation \cite{Papaioannou_AppliedEnergy_GUIDE_2025}. These developments suggest that LLM-based assistants could become key enablers of scalable, high-quality customer interaction in TOU/DR roll-out, DER interconnection and resilience-oriented outage communication. However, general LLMs lack robust understanding of sector-specific terminology, procedures \cite{Huang_ACMCS_2025} and regulatory frameworks \cite{Farquhar_Nature_2024}, leading to factual errors and potential rule violations. Existing evaluation practices prioritise surface fluency and operational-efficiency metrics such as average handling time, while underweighting technical accuracy, regulatory compliance, perceived user experience, and impacts on TOU adherence \cite{Jorgensen_Energies_2025}, DR execution and interconnection compliance \cite{Enrich_JPubEcon_2024}. Without a domain-grounded, multidimensional evaluation tailored to EPM, it is difficult to judge whether LLM-based assistants truly deliver trustworthy, grid-supportive and auditable guidance for sustainable power systems, or merely provide superficially fluent but unreliable responses.

\begin{figure}
    \centering
    \includegraphics[width=1\linewidth]{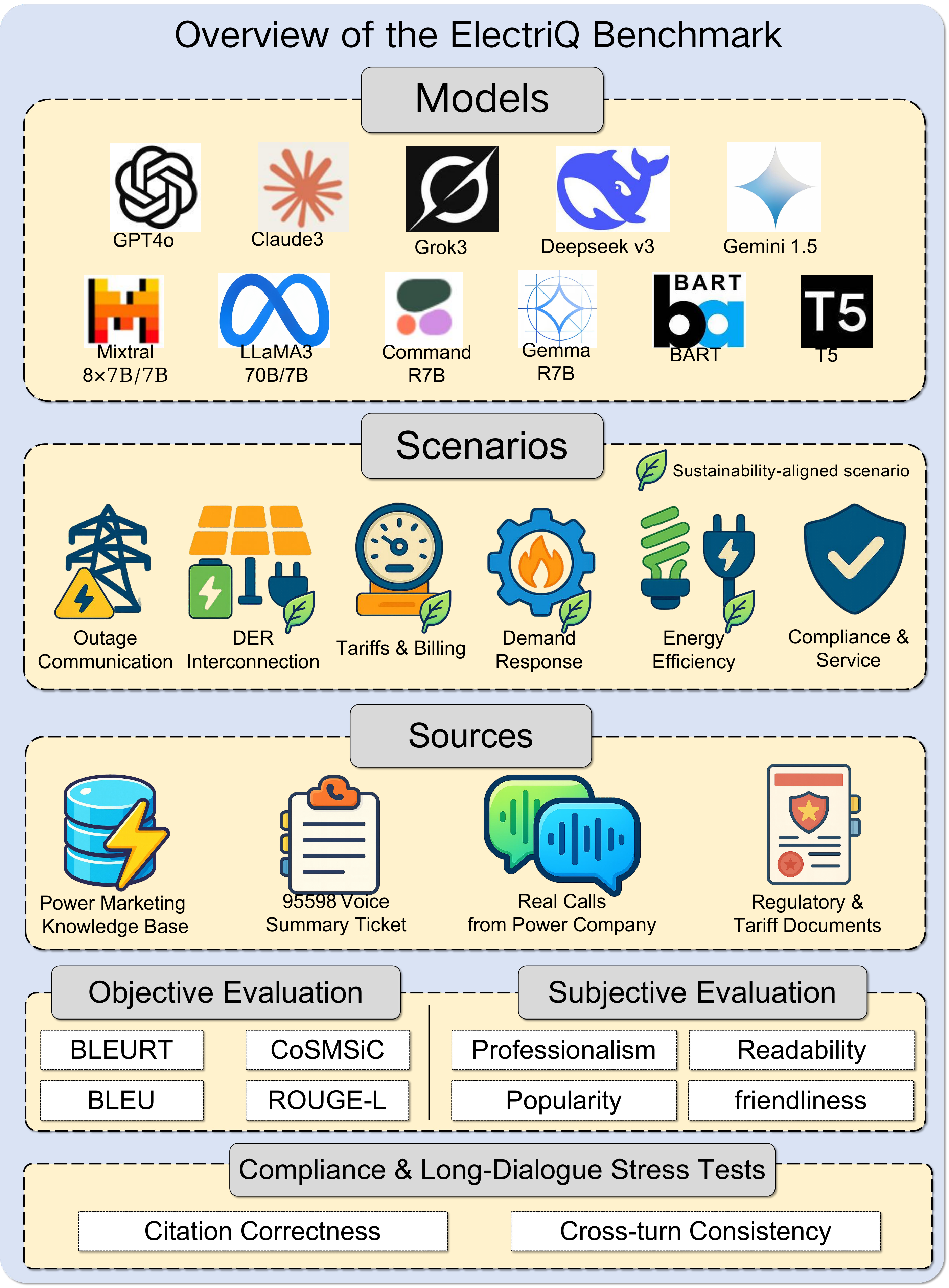}
    \caption{\textbf{Overview of the ElectriQ benchmark and evaluation. }
ElectriQ's dataset composition and scale, covered service scenarios and model spectrum, and unified evaluation with human and automatic metrics for EPM dialogues.}
    \label{fig2}
\end{figure}

To address these gaps at the customer-interface level, we present ElectriQ, a large-scale benchmark for evaluating LLMs in electric power marketing scenarios (Fig.~\ref{fig2}). ElectriQ covers six user-side service domains and 24 sub-scenarios that are central to TOU/DR implementation, DER interconnection and outage management, and includes over 550{,}000 multi-turn dialogues from power-service hotlines, 95598 work orders and logs, internal knowledge bases, and regulatory and tariff documents. We define a unified evaluation framework that combines standard automatic metrics with human ratings on multiple dimensions, and add two stress tests—Statutory Citation Correctness (SCC) and Long-Dialogue Consistency (LDC)---to probe regulatory citation, long-dialogue coherence and contractual stability. We further propose SEEK-RAG, a retrieval-augmented finetuning method that injects policy and domain knowledge during training and inference, improving the professionalism and compliance of model outputs in energy-marketing tasks. Experiments on 13 mainstream LLMs show that domain-aligned, knowledge-enhanced small models can match or surpass general-purpose LLMs at much lower computational cost, enabling more cost- and energy-efficient deployment of AI assistants in utility operations. In this way, ElectriQ provides a reproducible, auditable benchmark and a quantitative basis for advancing user-side intelligence as an integral component of demand-side management and sustainable-energy systems.

In summary, our contributions are threefold:
\begin{enumerate}
  \item We build ElectriQ, a large-scale dataset for EPM customer service (\textgreater{}550{,}000 multi\mbox{-}turn dialogues across six core domains and 24 sub-scenarios) that explicitly encode TOU/DR execution, DER interconnection and outage-management tasks in renewable-integrated power systems.
  \item We introduce a unified benchmark for EPM, defining multidimensional criteria such as professionalism, regulatory compliance and user experience, with standard metrics, open evaluation scripts and stress tests that capture contractual and regulatory robustness at the customer interface.
  \item We propose SEEK\mbox{-}RAG, a retrieval-augmented knowledge-augmentation method that improves professionalism and compliance, and evaluate 13 LLMs spanning sizes and architectures to inform cost- and energy-efficient model selection and configuration for utility-scale deployment.
\end{enumerate}
\section{Related Work}
Electricity provider management (EPM), exemplified by China’s ``95598'' customer-service system \cite{SGCC_95598_2025}, has standardised large-scale workflows for repairs, billing, and policy interpretation \cite{SGCC_Guide_2024}. Yet as user bases grow, business rules iterate, and regional heterogeneity widens, structural limits increasingly hinder alignment with sustainability goals. Prior work highlights inconsistent guidance on time-of-use (TOU) pricing and demand response (DR), which weakens demand-side flexibility \cite{Vasebi_2024_RSERR}; complex distributed energy resource (DER) interconnection rules, which invite divergent interpretations and delays \cite{DOE_i2X_Roadmap_2025}; shallow, weakly contextualised efficiency and EV-charging advice, which limits measurable savings \cite{Brambilla_2024_Informatics}; fragmented outage and restoration communication, which erodes grid resilience and trust \cite{DataCapable_2024_OutageJourney}; and siloed, asynchronously updated knowledge bases across channels, which make ``different answers to the same question'' more likely \cite{SGCC_2025_AI_Service}.

Motivated by these limitations, the power sector is exploring LLM-centric service paradigms. Recent work leverages long-context reasoning, retrieval-augmented generation (RAG), and tool orchestration for life-cycle sustainability assessment and energy-transition planning \cite{SustainLLM_2025_SETA}, personalised energy services and bill optimisation \cite{Arslan_2024_EB}, coupling with dispatch and forecasting to model load and charging behaviour in EV-rich systems \cite{Bienstock_2024_GridFM_arXiv}, and dialogue-based assistants for household energy efficiency and sustainable automation \cite{Varshney_2024_RAGTruth_ACL}. Overviews of ``grid foundation models'' outline a research agenda on data-driven, cross-task transfer in power systems, and domain-aligned dispatch models (e.g., GAIA) illustrate the benefits of combining domain alignment with retrieval and tool use for professional tasks \cite{Bienstock_2024_GridFM_arXiv}. Industrial pilots of human--AI collaboration in customer service similarly emphasise knowledge alignment and auditable processes \cite{SGCC_2025_AI_Service}.

Despite these advances, three gaps remain for frontline EPM. First, most studies emphasise planning, dispatch, or small-scale household prototypes, with little systematic, comparable evaluation anchored in core EPM processes such as pricing and settlement, DER interconnection, outage communication, and compliance appeals \cite{Yu_2024_RAG_Survey_arXiv}. Second, prevailing evaluations focus on fluency and generic accuracy rather than auditable, regulation-anchored metrics---especially citation correctness and cross-turn consistency---needed in regulated settings \cite{Zhang_2024_MT_Eval_EMNLP,Sirdeshmukh_2025_MultiChallenge_Findings}. Third, public data and protocols are scarce, and business-aligned KPIs tied to sustainability goals---such as TOU and DR implementation, DER interconnection guidance, and outage-communication transparency---lack unified benchmarks \cite{IEA_2024_EE_KeyFindings}. Our work addresses these gaps by introducing ElectriQ, an open, policy-verifiable benchmark aligned with sustainability objectives and grounded in real EPM dialogues.
\section{Task Setup}
We formulate EPM customer service as a regulated, multi-turn, policy-aligned dialogue generation task. Each instance consists of a user inquiry, optionally augmented with de-identified context (e.g., region), and a system reply. Model outputs must be actionable and auditable against applicable regulations, tariff terms, and interconnection rules. We evaluate all models under a unified protocol that combines subjective human ratings with two stress tests: statutory citation correctness and long-dialogue consistency.

\subsection{Evaluation metrics} \label{metrics}
Generic text metrics such as BLEU~\cite{10.3115/1073083.1073135} and ROUGE~\cite{lin-2004-rouge}, originally designed for translation and summarization, do not capture the domain constraints or user-experience requirements of EPM customer service. Accordingly, we define four EPM-oriented subjective metrics, grounded in a literature review, cross-sector standards (education \cite{tiffin2011evaluating}, healthcare \cite{ghafourian2023readability}, construction engineering \cite{karunaratne2022review}), analysis of 12,450 anonymized complaints (2022–2023), a 30-expert Delphi study, and user surveys:

\textbf{Professionalism}: regulatory and procedural correctness, including accurate terms and parameters; clause-consistent pricing, billing, restoration and suspension, and DER interconnection; region- and version-specific information; and internal logical consistency.

\textbf{Clarity \& Organization}: plain-language explanations of key terms (e.g., TOU/DR); clear structure (paragraphs /bullets /steps); coherent sentences; consistent units/labels; minimal ambiguity and redundancy.

\textbf{Actionability \& Completeness}: task-ready guidance covering required materials/forms, steps and order, channels/entry points, time windows/deadlines, costs/pricing rules, scope and uncertainties, plus clarification or human fallback when information is insufficient.

\textbf{Empathy \& Helpfulness}: respectful, reassuring tone; transparent expectations (uncertainties/updates); context-specific tips (e.g., outage safety); alternative contacts and accessible options; and smooth escalation when needed.

Each dimension is scored on a standardized 1–5 scale; detailed rubrics and representative 1- and 5-point examples are provided in Supplementary A.1. To improve objectivity and reproducibility, we also report four automatic metrics (BLEURT, CoSMSiC, BLEU, ROUGE) that capture semantic consistency, contextual coherence, lexical overlap, and information coverage. Together with human ratings, these yield a compact, multi-dimensional evaluation of model performance.
\subsection{Compliance \& Long-Dialogue Stress Tests}\label{stress}
To capture compliance risks that generic metrics miss, we introduce two stress tests: Statutory Citation Correctness (SCC) and Long-Dialogue Consistency (LDC). SCC checks whether responses contain verifiable legal citations (document title, version, clause) and whether key statements and numerical values match the cited clauses. LDC probes state tracking over 8--12 turns and requires consistent updates to references and commitments when the region or policy version changes. Both tests use binary checklists aggregated into 0--1 scores (reported as percentages), macro-averaged across six scenarios and 24 sub-tasks, with separate reporting for sustainability-aligned subsets. All items are double-annotated, and we report inter-annotator agreement.

\subsubsection{Statutory Citation Correctness (SCC)}
SCC stress-tests clause alignment and traceability in regulated EPM services. Under a ``content-alignment-first, citation-second'' principle, responses need not quote regulations verbatim, but they must be consistent with current clauses on key elements such as TOU windows, eligible user categories, required materials, and processing timelines. To support auditability, answers are encouraged to include concise source cues (for example, year and article or section), so SCC jointly evaluates factual alignment and source traceability across tariffs and settlement, distributed-energy interconnection, and related sustainability scenarios.

For each sample, we identify the scenario and key elements, then retrieve candidate clauses from a knowledge base indexed by document, version, and article; details of the retrieval and evidence-selection scoring are given in the Supplementary Material. We allow reasonable paraphrases and treat semantically equivalent entities as matches, while each scenario defines a set of required fields and missing fields are scored as 0. Closed-book and SEEK-RAG settings share the same evaluator to quantify compliance gains from retrieval augmentation.

The SCC score is defined as:
\begin{equation}
S_{\mathrm{CC}} = 100 \times \bigl( 0.7 \times C_{A} + 0.2 \times S_{A} + 0.1 \times R_{T} \bigr)
\end{equation}
where $C_{A}$ is the average accuracy over required fields (0--1), $S_{A}$ measures consistency on region, version, and validity period (0--1), and $R_{T}$ indicates whether the answer includes at least two citation elements (title, year, article or section; 1 if satisfied, 0 otherwise). To penalise ``citing without aligning'', scores are capped at 60 when $C_{A} < 0.5$.Formal definitions and a worked example are provided in the Supplementary Material (Section A.2).
\subsubsection{Cross-Turn Consistency (LDC)}
LDC evaluates whether models maintain stable process reasoning and state tracking over long multi-turn dialogues. It covers three capabilities: slot consistency with evidence-based updates as the dialogue progresses, valid workflow transitions under a predefined finite-state machine (FSM), and continuity or justified changes of sources and applicability across turns.

For each 8--12 turn dialogue, annotators label slot values and workflow states at every turn, check that changes to key fields are supported by new information, and verify that shifts in region or policy version are reflected in updated references and applicability. Dialogue-level scores are then averaged within each scenario.

We compute a single LDC score:
\begin{equation}
S_{\mathrm{LDC}} = 100 \times \bigl( 0.7 \times S_{\mathrm{CS}} + 0.2 \times S_{\mathrm{TV}} + 0.1 \times R_{\mathrm{CI}} \bigr)
\end{equation}
where $S_{\mathrm{CS}}$ measures per-turn slot agreement or justified updates, $S_{\mathrm{TV}}$ is the share of valid FSM transitions, and $R_{\mathrm{CI}}$ is the average rate of continuation or justified updates of sources and applicability. All components are normalised to $[0,1]$. Formal definitions and a worked example are provided in the Supplementary Material (Section A.3).
\section{Dataset Construction}\label{data}
\begin{figure*}
    \centering
    \includegraphics[width=1\linewidth]{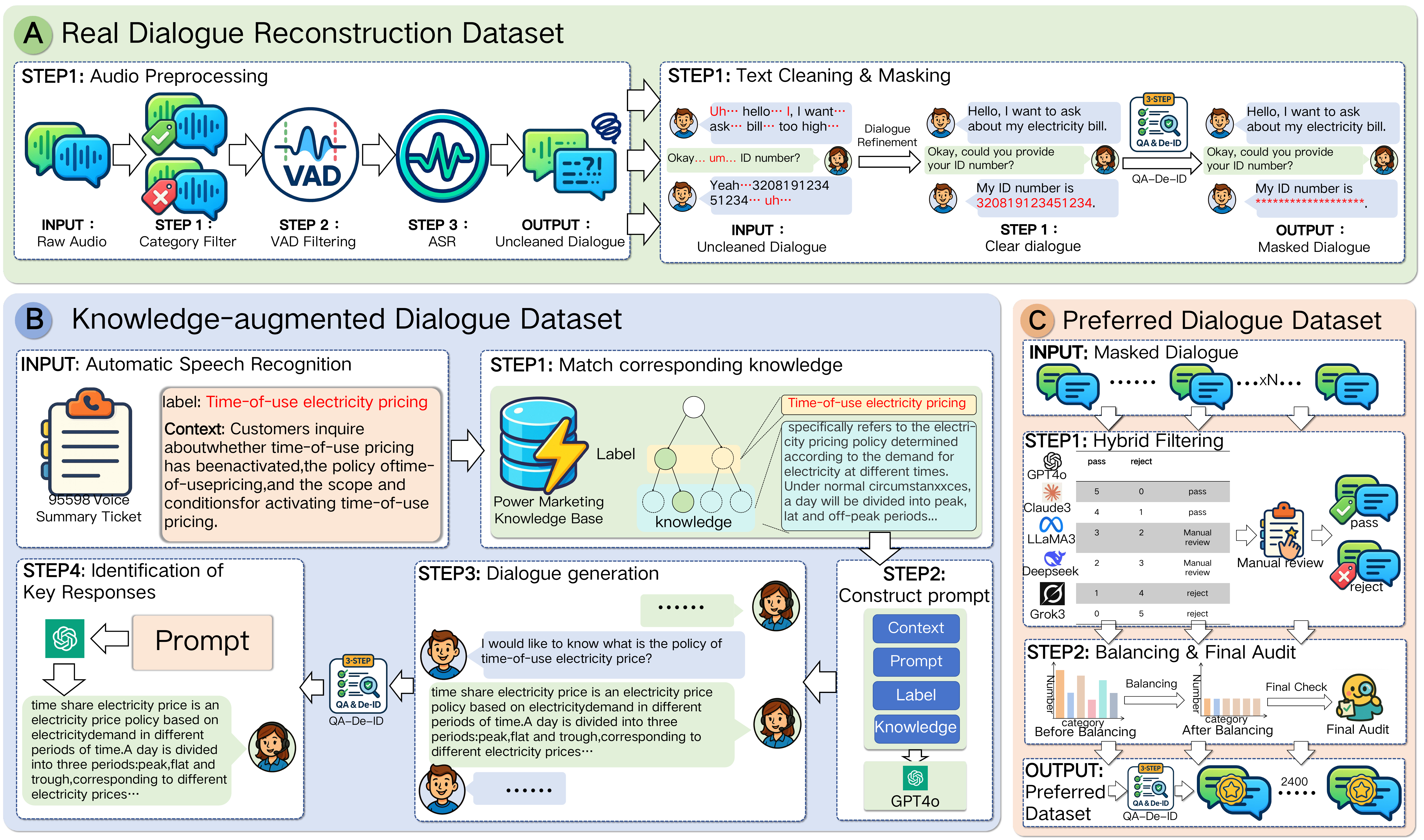}
    \caption{\textbf{Construction of the three ElectriQ subsets.}
    (A) 1.2M call logs $\rightarrow$ denoising, ASR, normalisation, de-identification $\rightarrow$ clean real multi-turn dialogues. 
    (B) Scenario-guided knowledge retrieval and prompted generation on aligned tickets and knowledge bases (training only, with the same de-ID and QA pipeline). 
    (C) Stratified sampling by complexity, urgency and scenario, followed by model- and human-based filtering with quota balancing, to obtain $\sim$2{,}400 preference-aligned samples.
    }
    \label{fig3}
\end{figure*}

ElectriQ assembles an EPM dialogue corpus aligned with sustainability and compliance. As shown in Fig.~\ref{fig3} and Table~\ref{ElectriQ Dataset Summary}, we combine customer-service audio logs, 95598 service tickets, internal knowledge bases, and regulatory/tariff documents into three complementary subsets: a Real Dialogue Reconstruction set built from denoised, transcribed and anonymised call recordings; a Knowledge-Augmented set that aligns tickets and knowledge-base entries with structured regulatory and tariff corpora and uses retrieval-augmented generation to synthesise clause-traceable dialogues; and a Preference-Aligned set selected by hybrid model--human screening and quota balancing. All samples are mapped to six service domains and 24 subdomains, with sustainability-critical scenarios (TOU/DR, DER interconnection, outage/restoration, efficiency and EV support) explicitly tagged for hierarchical analysis. The corpus contains over 550,000 multi-turn dialogues.

\begin{table*}[h]
\caption{\label{ElectriQ Dataset Summary} \textbf{ElectriQ dataset subsets.} Source, size, and coverage of domain knowledge (D), behaviour patterns (B), dialogue ability (L), and preference alignment (P).}
\centering
\renewcommand{\arraystretch}{1.2}
\scriptsize
\begin{tabular}{p{4.4cm} p{4cm} p{0.9cm} cccc}
    \toprule
    \textbf{Dataset} & \textbf{Source} & \textbf{Size} & \textbf{D} & \textbf{B} & \textbf{L} & \textbf{P} \\
    \midrule
    Real Dialogue Reconstruction Dataset & Real Calls
from Power Company & 188.4K & \checkmark & \checkmark & \checkmark & \ding{55} \\
    Knowledge-augmented Dialogue Dataset & 95598 Voice
Summary Ticket & 361.4K & \checkmark & \ding{55} & \ding{55} & \ding{55} \\
    Preferred Dialogue Dataset & Hybrid Screening & 2.4K & \checkmark & \checkmark & \checkmark & \checkmark \\
    \bottomrule
\end{tabular}

\vspace{2mm}
\raggedright
\scriptsize
D: Domain Knowledge; B: Behaviour Pattern; L: Dialogue Ability; P: Human Preference. 
\checkmark: contains ability; \ding{55}: does not contain ability.
\end{table*}

\subsection{Real Dialogue Reconstruction Dataset}
This subset is built from 1,200,000 customer-service call recordings (Table~\ref{ElectriQ Dataset Summary}; Fig.~\ref{fig3}(a)) via a two-stage pipeline for speech filtering and text reconstruction. Filtering first screens calls by business category to remove non-EPM and non-dialogue audio (e.g., system tests and back-end dispatch). Voice-activity detection and rule-based checks then keep only recordings with at least 80\% effective speech and 30~s or more of valid duration. The remaining 818,000 calls are transcribed by an ASR system adapted to power-sector terminology with an industry lexicon and custom dictionary, plus light normalisation of punctuation, numerals and time expressions to reduce domain-specific errors. Text reconstruction performs speaker separation and turn-boundary alignment to recover the agent--user multi-turn structure, followed by LLM-assisted normalisation (removing fillers and standardising formatting) to improve readability without changing factual content.

Quality and compliance are enforced through a three-step process (Supplementary Section B.1): automatic rule-based and PII detection, sampled human review with turn-level checks, and expert adjudication of borderline cases. The resulting corpus is a clean Real Dialogue Reconstruction Dataset with turn-level timestamps, speaker labels and preliminary scenario tags, usable for downstream training and evaluation and as input to the long-dialogue consistency stress test in Section~\ref{stress}.

\subsection{Knowledge-augmented Dialogue Dataset}
To broaden coverage, especially for sustainability-related cases, we construct retrieval-augmented (RAG) dialogues from the public 95598 corpus (around 1M requests) with explicit regulatory grounding. The original 95598 entries are single-turn records without multi-turn structure or clause-linked action details, so they are not suitable for direct training.

We align each request with EPM knowledge bases and regulatory/tariff documents (Fig.~\ref{fig3}(b)), use hybrid retrieval (BM25 plus vector) to collect evidence, and assemble structured prompts from labels and context. GPT-4o then generates policy-aligned, clause-anchored multi-turn dialogues, with clause metadata (document, year, section, region and validity) recorded, under safeguards that forbid fabricated clauses and constrain citation format, dialogue length and role/slot structure. Sustainability-focused scenarios are explicitly tagged and rebalanced, and we reuse the same three-stage quality and compliance review as for the Real Dialogue Reconstruction Dataset (Supplementary Section B.1). Validated dialogues are finally parsed by GPT-4o to extract key service replies and form high-quality training pairs. The Knowledge-augmented Dialogue Dataset is used only for training; evaluation relies on authentic dialogues and curated samples.

\subsection{Preferred Dialogue Dataset }
Using the cleaned, de-identified pool from the Real Dialogue Reconstruction Dataset, we build a Preference-Aligned Dialogue Dataset via the two-stage pipeline in Fig.~\ref{fig3}(c) to provide high-precision samples for alignment and evaluation. We first stratify dialogues by service complexity and issue urgency. A panel of review models then scores each candidate on four subjective dimensions and two objective metrics (SCC, LDC); high-scoring dialogues under a multi-model voting rule are kept (Supplementary Section B.1), low-scoring ones are discarded, and borderline cases receive rapid double review with senior adjudication. This stage only performs selection (no editing) and enforces SCC~$\geq$~80 and LDC~$\geq$~0.70.

From the retained pool, quota balancing produces about 2,400 dialogues (around 400 per service domain), guarantees minimum coverage of sustainability-related scenarios, and stratifies by dialogue length (short, medium, long) and speaking duration. A final pass checks label consistency, scenario alignment and text quality, reapplies the three-step quality-control and de-identification procedure (Supplementary Section B.1), and removes near-duplicates against the evaluation set to prevent leakage. The resulting Preference-Aligned Dialogue Dataset is stylistically consistent, high quality and auditable.

\subsection{Overall Dataset Analysis}
\begin{figure*}
    \centering
    \includegraphics[width=1\linewidth]{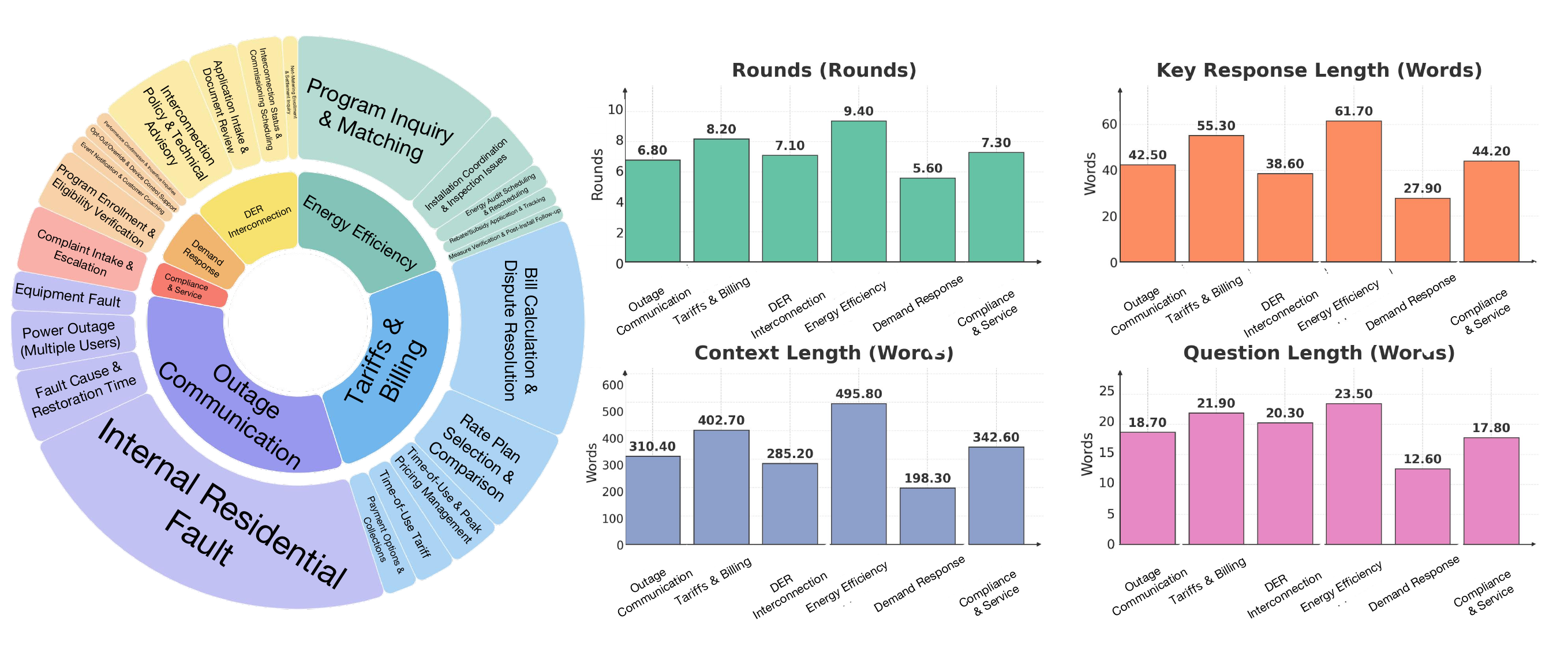}
    \caption{\textbf{ElectriQ taxonomy and per-category distributions.}
    Sunburst view of the ElectriQ taxonomy (six primary categories from >60 labels) and per-category dialogue-length statistics.
    }
    \label{fig:placeholder}
\end{figure*}
We analyse the ElectriQ corpus at a macro level to check category coverage and stability. As shown in Fig.~\ref{fig:placeholder}, over sixty fine-grained business labels are grouped into six primary and about twenty-four secondary categories. This two-level taxonomy reduces overlap and ambiguity while preserving key EPM scenarios, enabling reuse and comparable evaluation. Sustainability-related subsets (TOU/DR guidance, DER interconnection, outage/restoration, efficiency and EV support) are explicitly flagged for hierarchical analysis and ablations.

The four distributions on the right summarise, by category, the number of turns, key-response length, context length and question length. Overall ranges are controlled: mean turns 5.6--9.4, key responses about 28--62 tokens, contexts about 200--500 tokens, and questions 13--24 tokens. Categories with more procedural or technical content (for example, service expansion/interconnection and faults/outages) show longer contexts and replies, whereas general inquiries are shorter. We see no abnormal peaks or edge effects due to data construction, suggesting that cleaning and reconstruction did not introduce systematic truncation.

All samples retain metadata such as region, tariff version and validity period, supporting the regulatory-citation and cross-turn-consistency stress tests in Section~\ref{stress}. Together with the unified QA and de-identification pipeline, this yields a clearly structured, semantically well-covered and low-noise corpus that provides a reliable, auditable basis for modelling and evaluation across categories and sustainability-related subsets.
\section{Experimental Methodology}

Building on the ElectriQ dataset in Section~\ref{data}, this section introduces the ElectriQ-LLM training strategy and the SEEK-RAG mechanism, targeting professional, auditable and practically useful responses for EPM workflows such as TOU promotion, DR execution and DER interconnection.
\subsection{Model Training Strategy}
\begin{figure}
    \centering
    \includegraphics[width=1\linewidth]{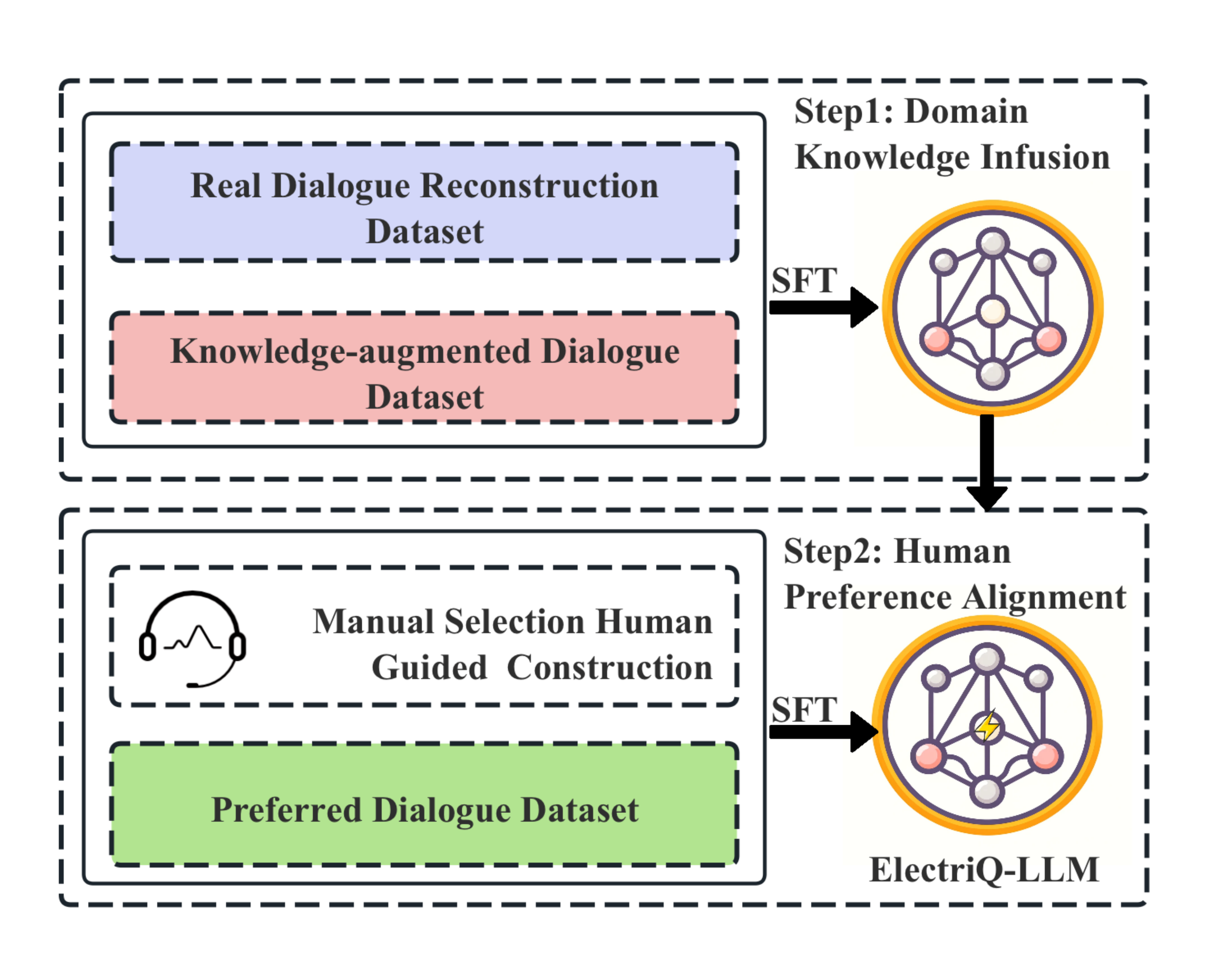}
    \caption{\textbf{Two-stage training of ElectriQ-LLM.}
    Stage~1 injects domain knowledge using large-scale EPM dialogues; Stage~2 aligns responses to human preferences on a 2,400-sample subset to improve professionalism and user experience.
    }
    \label{Two stage training process of ElectriQ-LLM}
\end{figure}

To balance domain professionalism, compliance and actionability, ElectriQ-LLM is trained with a two-stage strategy (Fig.~\ref{Two stage training process of ElectriQ-LLM}), with sustainability-related subsets and compliance goals shaping both sampling and preference objectives.

\textbf{Stage One: Domain Knowledge Infusion.}
Stage~1 fine-tunes the base model on large-scale Electric Power Marketing (EPM) dialogues using the Real Dialogue Reconstruction and Knowledge-Augmented Dialogue datasets. Weighted sampling and a curriculum schedule start from general scenarios and gradually increase sustainability-focused samples, so the model first learns generic conversational patterns before specialising in load shifting, renewable integration and resilience-oriented user support. To reduce compute and energy cost for large-scale utility deployments, we adopt QLoRA fine-tuning in this stage.

\textbf{Stage Two: Human Preference Alignment.}
Stage~2 aligns the model on the 2,400-sample Preference-Aligned Dialogue Dataset to improve user experience, compliance and executability. Preferences combine the four subjective dimensions in Section~\ref{metrics} with the objective metrics SCC and LDC. Training uses supervised fine-tuning (SFT) followed by direct preference optimisation (DPO): SFT stabilises response style, and DPO steers the model towards behaviours such as actionable guidance, appropriate citation and proper escalation. The resulting ElectriQ-LLM produces answers that are easy to understand, provide stepwise, operational recommendations with concise source cues and clearly signal when escalation to human agents is needed. Overall, the two-stage design balances performance, compliance and energy efficiency and fits high-volume utility services such as TOU tariff rollout, DR participation support and DER interconnection consultation.
\subsection{SEEK-RAG: Scene- and Keyword-Guided Retrieval-Augmented Generation}

\begin{figure}
    \centering
    \includegraphics[width=1\linewidth]{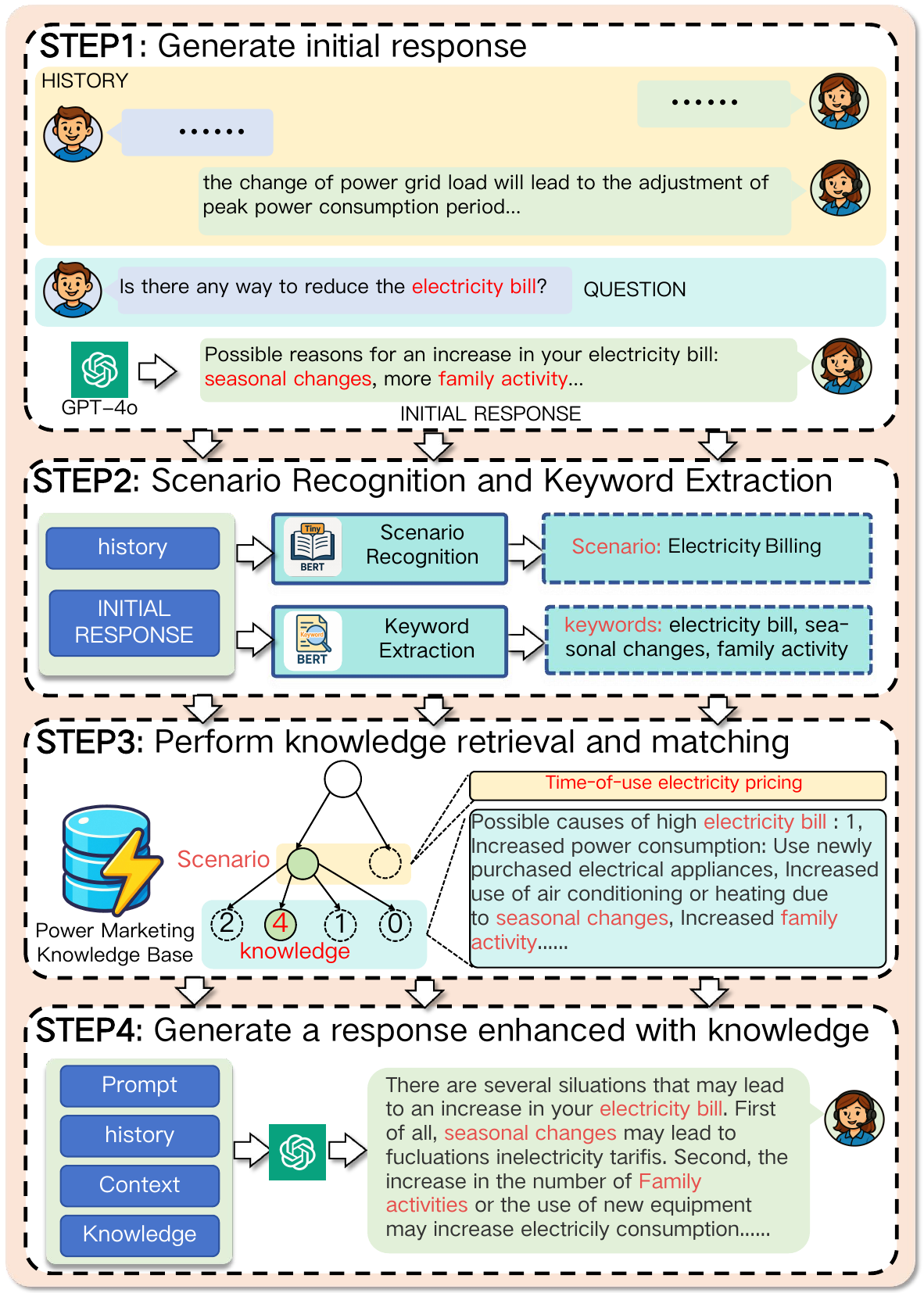}
    \caption{\textbf{SEEK-RAG: Scene- and keyword-guided retrieval-augmented generation.}
    Four steps: initial response, scene/keyword extraction, scenario-filtered retrieval with keyword coverage, and knowledge-enhanced generation.
    }
    \label{SEEK-RAG}
\end{figure}

To improve accuracy, interpretability and actionability in EPM customer service, we employ SEEK-RAG, a four-step pipeline (Fig.~\ref{SEEK-RAG}) that uses TinyBERT for scene recognition and a BERT-based keyword extractor to guide LLM retrieval. Let \(H\) be the dialogue history, \(Q\) the current user query, \(P\) a task prompt and \(\mathcal{K}\) the domain knowledge base.

\paragraph{Step 1: Initial response.}
The LLM first takes \(P\), \(H\) and \(Q\) and outputs an initial reply \(R_0\):
\begin{equation}
R_0 = \mathrm{LLM}(P, H, Q).
\end{equation}

\paragraph{Step 2: Scene and keyword extraction.}
To avoid long-history bias, we use only \((Q, R_0)\). TinyBERT predicts scene labels \(S\), and the keyword extractor returns salient terms \(K_{\text{ext}}\):
\begin{equation}
S=\mathrm{TinyBERT}_{\text{scene}}(Q, R_0), \qquad
K_{\text{ext}}=\mathrm{BERT}_{\text{key}}(Q, R_0).
\end{equation}

\paragraph{Step 3: Knowledge retrieval and matching.}
Using the predicted scenes, we restrict the knowledge base to the union of scenario-specific sublibraries:
\begin{equation}
\mathcal{K}_S=\bigcup_{s\in S}\mathrm{SubLibrary}(\mathcal{K}, s).
\end{equation}
For each candidate item \(k_j\in\mathcal{K}_S\), we define a keyword-coverage score
\begin{equation}
\mathrm{score}(k_j)=\sum_{w\in K_{\text{ext}}}\mathbf{1}\!\left[w\in k_j\right],
\end{equation}
and select the best-matched knowledge point:
\begin{equation}
K_{\text{match}}=\underset{k_j\in \mathcal{K}_S}{\arg\max}\ \mathrm{score}(k_j).
\end{equation}

\paragraph{Step 4: Knowledge-enhanced response.}
Finally, the LLM conditions on \(P\), \(H\), \(Q\), \(R_0\) and \(K_{\text{match}}\) to produce the enhanced response:
\begin{equation}
R_{\text{final}}=\mathrm{LLM}\!\big(P, H, Q, R_0, K_{\text{match}}\big).
\end{equation}

In summary, SEEK-RAG combines scene and keyword cues with a simple coverage-based matching criterion to inject structured domain knowledge into generation. Under frequently updated tariffs and interconnection rules, this helps track regional and version differences, lowers the risk of misguiding users and keeps the system easy to audit in utility settings.
\section{Experimental Setups}
\subsection{Dataset}
We use the ElectriQ corpus (Sec.~\ref{data}), treating the Knowledge Augmented subset as training-only and building validation/test sets only from real and preference-aligned dialogues with cross-session deduplication. Unless noted otherwise, we adopt an 80/10/10 stratified train–val–test split, train on the full Knowledge-Augmented subset with random seed 42, and include four annotated stratified test subsets for sustainability analysis. For compliance evaluation, the retrieval index is frozen before testing and isolated from training-time regulatory versions; de-identification and quality control follow Sec.~\ref{data}.

\subsection{Baselines}
Because no public EPM-specific LLMs exist, we benchmark general-purpose models in four size tiers. \textbf{Extra-large:} GPT-4o, Claude~3, Gemini~1.5, Grok~3, DeepSeek-V3. \textbf{Large:} Llama~3 70B, Mixtral 8$\times$7B. \textbf{Medium:} Llama~3 8B, Mistral 7B, CommandR 7B, Gemma 7B. \textbf{Small:} mT5-base, BART-large.

\subsection{Evaluation Metrics}
We adopt a hybrid framework with automatic and human metrics. Automatic metrics (BLEURT, CoSMSiC, BLEU, ROUGE-L) assess semantic consistency, lexical similarity and information coverage. Human evaluation uses four dimensions—professionalism, clarity/organization, executability/completeness and empathy/usefulness—to capture both compliance/actionability and user experience. We also report two stress tests, SCC and LDC. All models use identical prompts and decoding budgets, and results include 95\% confidence intervals and significance tests.
\subsection{Implementation Details}
We performed parameter-efficient fine-tuning of mT5-base, BART-large, and several 7B-scale models on three NVIDIA RTX~3090 GPUs. For mT5-base and BART-large, the learning rate was set to 5e$^{-5}$, with 20 training epochs, a batch size of 12, \texttt{bf16} precision, and the AdamW optimizer with a cosine learning rate schedule. The context window size was 4k tokens. For the 7B models, we applied QLoRA fine-tuning with rank set to 64, $\alpha = 16$, and dropout = 0.05. The learning rate was 1e$^{-4}$, training was conducted for 20 epochs with a batch size of 10, and \texttt{bf16} precision was used. For the TinyBERT model used in scenario classification, the learning rate was set to 3e$^{-5}$, with 15 epochs and a batch size of 16. For the BERT-key model used for keyword extraction, the learning rate was set to 2e$^{-5}$, with 25 epochs and a batch size of 8.

Before evaluation, the RAG index was frozen. We adopted a hybrid scoring strategy combining BM25 ($k_1 = 1.2$, $b = 0.75$) with keyword coverage, setting the top-$k$ value to 5. The unified decoding configuration included a temperature of 0.2, a beam width of 4, and a maximum generation length of 512. A total of 1,000 test samples were stratified from real dialogues, covering six main categories, twenty-four subcategories, and the four sustainability-related subsets, to ensure sample diversity and representativeness. All experiments were conducted with fixed random seeds and identical decoding budgets.
\section{Experiments}
To evaluate ElectriQ and our methods under a unified protocol, we study five research questions (RQs):
\begin{itemize}
    \item \textbf{RQ1:} Performance of LLMs of different scales and architectures on ElectriQ overall and on the four sustainability-focused subsets.
    \item \textbf{RQ2:} Correlations among the four human-rated dimensions and how they change after SFT and SEEK-RAG.
    \item \textbf{RQ3:} Contributions of SFT and SEEK-RAG to performance gains across model sizes, and comparison with Zero-shot-CoT.
    \item \textbf{RQ4:} Effects of model scale on SCC and LDC, and whether SFT and SEEK-RAG significantly improve these metrics, especially in sustainability-related scenarios.
    \item \textbf{RQ5:} Impact of real vs.\ synthetic dialogues on ElectriQ results and the generalisability of our methods to other power subdomains (e.g., photovoltaics, hydropower, substations).
\end{itemize}
\subsection{Overall and sustainability-focused performance (RQ1)}
Under the unified protocol in Supplementary Section B.2, each model generates a key response that is scored by five frontier LLMs (GPT-4o, Claude~3, Gemini~1.5, DeepSeek-v3, Grok~3). Five power-marketing experts re-score 500 samples, yielding a Cohen’s $\kappa$ of 0.73 with the LLM panel. We apply this protocol before and after ElectriQ-style adaptation (SFT + SEEK-RAG); Table~\ref{tab:sft-results} reports post-adaptation human scores, and Supplementary C.1 report baselines and automatic metrics.

\begin{figure}
    \centering
    \includegraphics[width=1\linewidth]{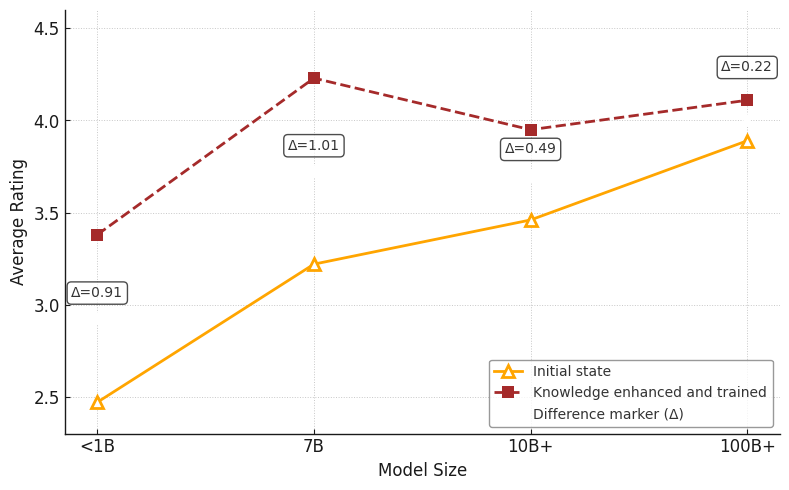}
    \caption{\textbf{Overall Subjective Scores vs. Model Scale.}
    English caption: Scores grow with size but show diminishing returns beyond ~10B; adding SFT+SEEK-RAG shifts the curve upward, with the largest gains for small and midsize models (e.g., ~4.2 for 7B).
    }
    \label{Different Volume LLMs Scores}
\end{figure}
Human ratings increase with model scale but show diminishing returns beyond $\sim$10B parameters. Before adaptation, averages are roughly 2.5 for sub-1B models, 3.2 for 7B, 3.4--3.5 for 10B+ and 3.9 for 100B+ models. After SFT + SEEK-RAG, the curve in Fig.~\ref{Different Volume LLMs Scores} shifts upward, with mean gains of about +0.9 (sub-1B), +1.0 (7B), +0.5 (10B+) and +0.2 (100B+). The 7B bucket reaches $\approx$4.2, essentially overlapping ultra-large models ($\approx$4.1) while using far less compute. Gains are largest in professionalism, actionability/completeness and empathy/helpfulness, with clarity/organization also improving. For example, LLaMA3-8B rises from $\sim$3.3 to 4.31 and Mistral-7B from $\sim$3.2 to 4.30, approaching LLaMA3-70B and GPT-4o. Ultra-large models, by contrast, see modest gains from SEEK-RAG alone. Overall, ElectriQ-style domain adaptation and retrieval augmentation let mid-sized models approach frontier-level service quality at a fraction of computational and energy cost.
\begin{table}
\centering
\scriptsize
\caption{\label{tab:sft-results} \textbf{Subjective evaluation after SFT and SEEK-RAG.} Values in parentheses denote absolute and relative gains over baseline.}
\label{tab:auto-metrics}
\begin{tabular}{@{}lcccccc@{}}
\toprule
\rotatebox{90}{} &Model         & BLEURT & CoSMSiC & BLEU  & ROUGE‑L \\
\midrule
\multirow{13}{*}{\rotatebox{90}{\textbf{Original }}} & 
GPT4o         & 0.839 & 0.889 & 43.4 & 45.7 \\
&Claude3       & \textbf{0.894} & \underline{0.909} & 42.9 & 46.5 \\
&Gemini 1.5    & 0.861 & 0.875 & 41.5 & 42.7 \\
&\textbf{Grok 3}        & \underline{0.892} & \textbf{0.919} &\textbf{ 43.6} & \textbf{47.6} \\
&Deepseek-v3   & 0.876 & 0.869 & \underline{43.5} & \underline{46.4} \\
&LLaMA3-70B    & 0.713 & 0.731 & 33.2 & 37.3 \\
&Mixtral 8x7B  & 0.792 & 0.813 & 38.2 & 40.1 \\
&LLaMA3-8B     & 0.718 & 0.764 & 35.5 & 36.7 \\
&Mistral-7B    & 0.690 & 0.713 & 32.8 & 34.6 \\
&CommandR-7B   & 0.685 & 0.720 & 30.4 & 35.3 \\
&Gemma-7B      & 0.680 & 0.691 & 30.0 & 34.8 \\
&BART-large    & 0.525 & 0.537 & 21.1 & 26.8 \\
&MT5-base      & 0.511 & 0.501 & 19.9 & 24.8 \\
    \hline
\multirow{13}{*}{\rotatebox{90}{\textbf{SFT-KE}}} & 
GPT4o         & 0.874 & 0.868 & 41.4 & 45.2 \\
&Claude3       & 0.892 & 0.873 & 43.1 & 44.7 \\
&Gemini 1.5    & 0.855 & 0.833 & 40.7 & 43.0 \\
&Grok 3        & 0.865 & 0.888 & 43.3 & 43.6 \\
&Deepseek-v3   & 0.876 & 0.856 & 42.4 & 44.3 \\
&LLaMA3-70B    & 0.776 & 0.791 & 37.8 & 39.9 \\
&Mixtral 8x7B  & 0.724 & 0.720 & 34.7 & 37.1 \\
&LLaMA3-8b     & \textbf{0.946} & \underline{0.945} & \underline{44.6} & \underline{46.8} \\
&\textbf{Mistral-7B}   & \underline{0.940} & \textbf{0.959} & \textbf{47.1} & \textbf{46.9} \\
&CommandR-7B   & 0.906 & 0.911 & 42.8 & 45.6 \\
&Gemma-7B      & 0.864 & 0.873 & 42.5 & 45.0 \\
&BART-large    & 0.521 & 0.530 & 24.5 & 29.1 \\
&MT5-base      & 0.519 & 0.535 & 24.3 & 27.8 \\
\bottomrule
\end{tabular}
\end{table}
\begin{figure*}
    \centering
    \includegraphics[width=1\linewidth]{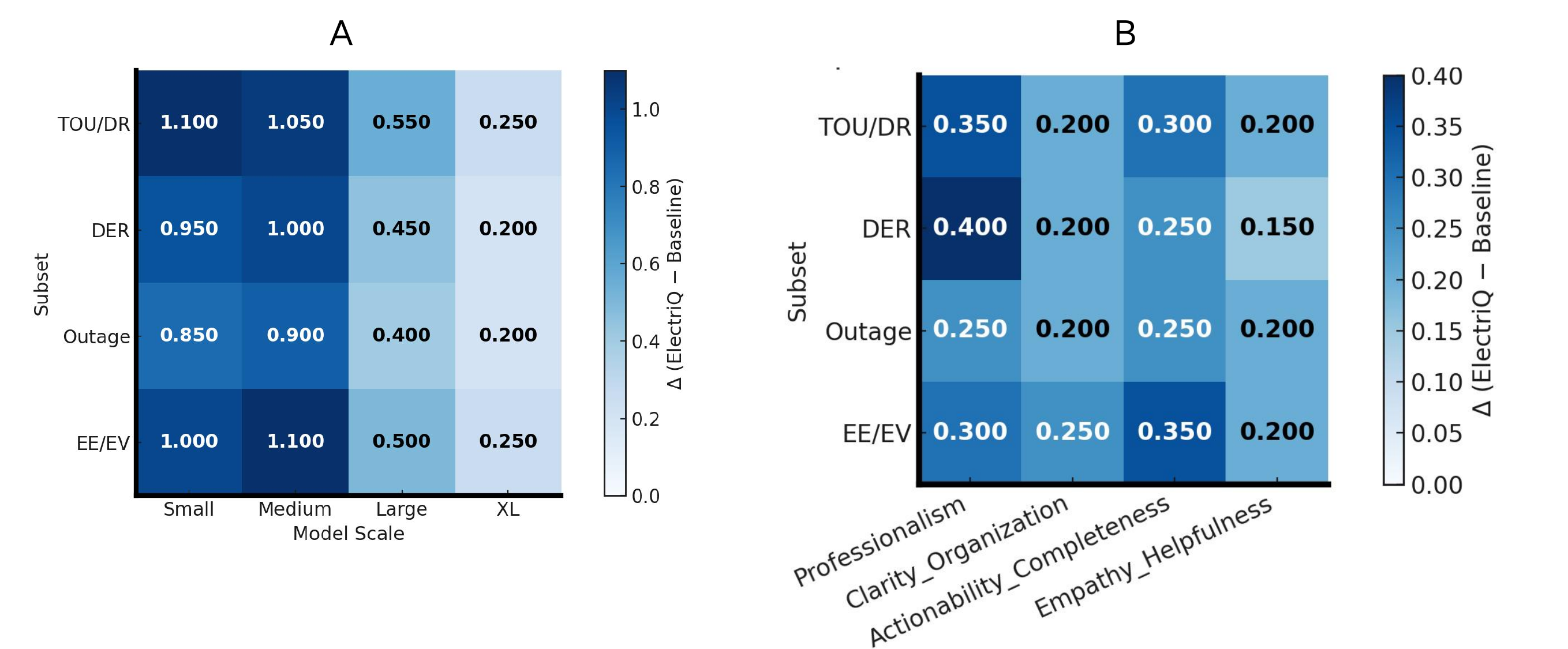}
    \caption{\textbf{Gains on sustainability subsets.}
(A) Macro-average human-rating gain $\Delta$ across subsets and model scales.
(B) For 7B models, per-dimension gains across subsets.
Color encodes $\Delta$; cell labels show values.}
\label{fig:rq2_heatmaps}
\end{figure*}

We next examine four sustainability-focused subsets: TOU/DR guidance, distributed-energy interconnection, efficiency and EV support, and outage/restoration communication. Using the same protocol and grouping models into four size buckets (sub-1B, 7B, 10B+, 100B+), Fig.~\ref{fig:rq2_heatmaps} shows patterns that largely mirror the full corpus: SFT + SEEK-RAG yields the biggest gains for small and medium models, especially 7B, whose average scores on all four subsets become nearly indistinguishable from those of 100B+ models. BLEURT, CoSMSiC, ROUGE-L and BLEU also improve consistently (Supplementary section C.1), indicating joint gains in semantic alignment, information coverage and readability, while non-adapted ultra-large models start high but have limited headroom.

At the scenario level, the largest gains appear in DER interconnection and outage/restoration, which involve process-intensive workflows. Here, SEEK-RAG strengthens use of procedural knowledge and key operational elements such as material checklists, deadlines, step sequences and communication templates. TOU/DR and efficiency/EV subsets show more balanced gains: terminology is simplified, recommendations become more executable and complete, and key information is presented more clearly. Overall, the sustainability subsets make ElectriQ more sensitive to decarbonisation-related tasks, and the results show that SFT + SEEK-RAG enable 7B-scale models to reach practically deployable service quality under realistic compute constraints, offering a cost-effective option for real-world utility customer service.
\subsection{Correlation Analysis of Metrics (RQ2)}
\begin{figure*}
    \centering
    \includegraphics[width=1\linewidth]{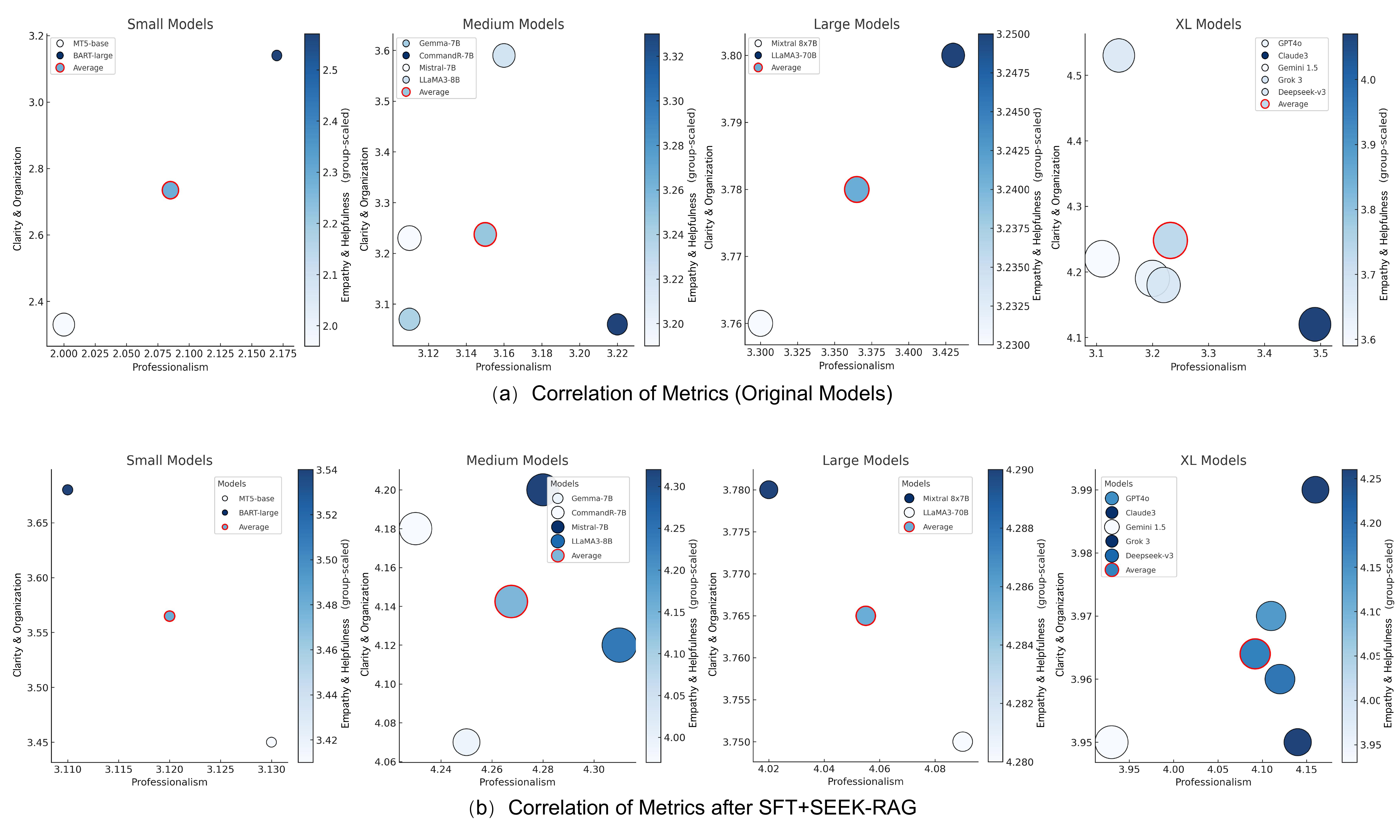}
    \caption{\textbf{Correlation of subjective metrics.}
    Professionalism (x), clarity/organization (y); colour encodes user-friendliness, area actionability/completeness.
    }
    \label{Metrics_Correlation_Comparison_Chart}
\end{figure*}

We analyse how the four subjective dimensions interact in sustainability-related scenarios using bubble plots (Fig.~\ref{Metrics_Correlation_Comparison_Chart}). On each plot, the horizontal axis is professionalism, the vertical axis clarity/organization, bubble colour user friendliness and bubble area actionability/completeness. Before adaptation, points already show a mild upper-right trend: higher professionalism tends to coincide with better clarity, user friendliness and slightly larger bubbles. After applying SFT and SEEK-RAG, most points move further into the upper-right region, especially for DER interconnection and outage communication, where retrieval-enhanced procedural steps and checklists yield noticeably larger bubbles. At very high professionalism levels, however, the curve flattens and clarity gains shrink, particularly in clause-dense TOU/DR and DER subsets. This suggests that simply adding detail can hurt readability, and that short glossaries, concise key-point summaries and standardised “what to do / how to do it” snippets are needed to keep highly professional replies both understandable and actionable.
\subsection{Ablation Study (RQ3)}
We compare five configurations for each model under the unified protocol in Section~4.5: (i) Base, (ii) SFT, (iii) SEEK-RAG, (iv) SFT+Zero-shot-CoT, and (v) SFT+SEEK-RAG. All share the same request set, prompts and decoding/retrieval settings. Table~\ref{Ablation Experiment Score Comparison} reports average human scores over the four subjective dimensions.

\begin{table}
\caption{\label{Ablation Experiment Score Comparison}\textbf{Ablation results.}
Average human scores for Base, SFT, SEEK-RAG, SFT+CoT and SFT+SEEK-RAG (higher is better).}
\centering
\scriptsize
\begin{tabular}{@{}lccccc@{}}
    \hline
    \textbf{Model} & \textbf{Original} & \textbf{SFT} & \textbf{RAG} & \textbf{SFT-COT} & \textbf{SFT-RAG} \\
    \hline
    GPT4o & \underline{3.87} & 3.87 & \underline{4.12} & 4.05 & 4.12 \\
    Claude3 & \textbf{3.98} & 3.98 & \textbf{4.13} & 4.04 & 4.13 \\
    Gemini 1.5 & 3.86 & 3.86 & 4.08 & 4.01 & 4.08 \\ 
    Grok 3 & 3.87 & 3.87 & 4.12 & 4.01 & 4.12 \\
    Deepseek-v3 & 3.88 & 3.88 & 4.13 & 4.01 & 4.13 \\
    LLaMA3-70B & 3.46 & 3.46 & 3.96 & 3.96 & 3.96 \\
    Mixtral 8x7B & 3.45 & 3.45 & 3.94 & 3.94 & 3.94 \\
    \textbf{LLaMA3-8B} & 3.32 & \underline{4.09} & 3.88 & \underline{4.14} & \textbf{4.31} \\
    Mistral-7B & 3.23 & \textbf{4.11} & 3.87 & \textbf{4.18} & \underline{4.30} \\
    CommandR-7B & 3.18 & 3.99 & 3.77 & 4.06 & 4.19 \\
    Gemma-7B & 3.15 & 4.03 & 3.93 & 4.05 & 4.10 \\
    BART-large & 2.55 & 3.53 & 2.77 & 3.42 & 3.42 \\
    MT5-base & 2.38 & 3.57 & 2.39 & 3.33 & 3.33 \\
    \hline
\end{tabular}
\end{table}

For sub-$10$B models, SFT is the main driver of improvement. Both 7B and sub-1B groups show large, stable gains of about $+0.8$--$1.2$ points (e.g., LLaMA3-8B from $3.32$ to $4.09$, Mistral-7B from $3.23$ to $4.11$, MT5-base/BART-large from $\approx 2.4$--$2.6$ to $\approx 3.5$). In contrast, SEEK-RAG's benefits grow with scale: 70B and mixture models gain about $+0.5$, and frontier models such as GPT-4o, Claude 3, Gemini 1.5, Grok 3, and DeepSeek-v3 gain roughly $+0.22$--$0.25$ by better exploiting retrieved evidence. Across all sizes, SFT+SEEK-RAG yields the best overall performance. For 7B models this combination is clearly super-additive: LLaMA3-8B reaches $4.31$ and Mistral-7B $4.30$, both surpassing either SFT or SEEK-RAG alone. For 70B and ultra-large models, SFT alone has limited effect, indicating that most gains at this scale stem from integrating retrieved knowledge. Under the same SFT setting, SEEK-RAG consistently outperforms zero-shot CoT.

On the four sustainability-focused subsets, SFT mainly stabilises terminology and procedural descriptions, improving interpretability and stepwise guidance. SEEK-RAG complements this by aligning tariff, regulatory and workflow parameters with current rules (e.g., peak/off-peak windows, eligible users, required materials and channels, processing timelines). In combination, 7B models reach average human scores of about 4.2 on these subsets, overlapping ultra-large models while using far fewer resources. Overall, ElectriQ’s dual-alignment strategy—task adaptation via SFT plus policy-grounded retrieval via SEEK-RAG—both boosts performance and yields stable, practically useful behaviour in user-facing, sustainability-critical scenarios.
\subsection{Compliance and Long-Dialogue Robustness (RQ4)}
We evaluate Regulatory Citation Correctness (SCC) and Long-Dialogue Consistency (LDC) on the ElectriQ test set, which contains only reconstructed and preference-aligned dialogues. All models share the same requests, prompts and decoding/retrieval hyperparameters. SFT is applied only to sub-1B and 7B models; larger models (10B, 70B, 100B+) are tested in Base and SEEK-RAG settings.

Table~\ref{tab:model_score_comparison} reports mean SCC scores (0--100) with 95\% confidence intervals. Across all models, SEEK-RAG is the main source of gain: adding retrieval raises SCC by about 10--11 points for frontier models, and by nearly 14 points for LLaMA3-70B and Mixtral 8$\times$7B (from roughly 62 to 76). For 7B models, combining SFT and SEEK-RAG yields the largest jumps: LLaMA3-8B and Mistral-7B rise from around 59 to about 80, and sub-1B models (mT5-base, BART-large) gain 17--18 points. SCC increases with model size but shows diminishing returns beyond 10B parameters, while domain-specific retrieval provides strong gains for both small and large models. Without access to regulatory content, scaling alone does not substantially improve citation accuracy. With SFT+SEEK-RAG, the 7B bucket reaches SCC scores near 80, matching the 77--78 range of 100B+ models at far lower computational cost, which is attractive for TOU/DR, DER and outage/restoration services that require verifiable, regulation-aligned advice.

\begin{table*}
\caption{\label{tab:model_score_comparison} 
\textbf{SCC scores (0--100).} Mean SCC under Base, SFT and SFT+SEEK-RAG; $\Delta$ columns show gains over the Base model (higher is better).}
\centering
\scriptsize
\begin{tabular}{@{}lccccc@{}}
    \hline
    \textbf{Model} & \textbf{Original} & \textbf{SFT} & \textbf{SFT+SEEK-RAG} & \textbf{$\Delta$ (SFT$-$Orig)} & \textbf{$\Delta$ (SFT$+$RAG$-$Orig)} \\
    \hline
    GPT-4o & 67.2 ± 1.3 & — & 78.3 ± 1.1 & — & +11.1 \\
    Claude 3 & 66.4 ± 1.4 & — & 77.2 ± 1.1 & — & +10.8 \\
    Gemini 1.5 & 66.9 ± 1.3 & — & 77.8 ± 1.1 & — & +10.9 \\
    DeepSeek-v3 & 66.1 ± 1.4 & — & 77.1 ± 1.2 & — & +11.0 \\
    Grok 3 & 67.4 ± 1.3 & — & 77.6 ± 1.1 & — & +10.2 \\
    LLaMA3-70B & 62.5 ± 1.4 & — & 76.2 ± 1.2 & — & +13.7 \\
    Mixtral 8$\times$7B & 62.1 ± 1.5 & — & 76.0 ± 1.2 & — & +13.9 \\
    \textbf{LLaMA3-8B} & 59.4 ± 1.5 & 66.8 ± 1.3 & 80.6 ± 1.1 & +7.4 & \textbf{+21.2} \\
    Mistral-7B & 58.6 ± 1.5 & 66.2 ± 1.3 & 80.1 ± 1.1 & +7.6 & +21.5 \\
    CommandR-7B & 58.1 ± 1.6 & 65.1 ± 1.4 & 79.5 ± 1.2 & +7.0 & +21.4 \\
    Gemma-7B & 58.8 ± 1.5 & 65.6 ± 1.3 & 80.4 ± 1.1 & +6.8 & +21.6 \\
    MT5-base & 51.6 ± 1.6 & 58.0 ± 1.4 & 69.7 ± 1.2 & +6.4 & +18.1 \\
    BART-large & 52.6 ± 1.6 & 59.1 ± 1.4 & 70.9 ± 1.2 & +6.5 & +18.3 \\
    \hline
\end{tabular}
\end{table*}

For LDC, we run Base, SFT, SEEK-RAG and SFT+SEEK-RAG for sub-1B and 7B models, and Base vs SEEK-RAG for 10B+ models (full scores in Supplementary Section C.2). LDC falls as dialogue length increases, but SEEK-RAG consistently improves performance across short, medium and long conversations, with the largest relative gains in 9--12 turns. Grouped by model scale, improvements follow the pattern small/medium $>$ large $>$ ultra-large: sub-1B and 7B models benefit most, 70B somewhat less and ultra-large models least. SFT+SEEK-RAG gives the best LDC for sub-1B and 7B models, narrowing the gap to ultra-large models and offering a better accuracy--efficiency trade-off. On sustainability-focused subsets with stricter regulations, retrieval reduces cross-turn contradictions and parameter drift, while SFT stabilises terminology and procedural descriptions. Together, higher SCC and LDC show that ElectriQ-style dual alignment enables models—especially 7B-scale ones—to deliver compliant, coherent and operationally reliable guidance for DR participation, TOU adoption, DER interconnection and outage-related support.
\subsection{Cross-source and Cross-subdomain Robustness (RQ5)}
We test whether synthetic data bias evaluation by comparing 13 models on 1{,}000 synthetic (S.D.) and 1{,}000 real (R.D.) instances under identical prompts and decoding/retrieval settings (full scores in Supplementary Section C.3). Average gaps between S.D. and R.D. are small: across models, relative differences stay within about $\pm 2.5\%$, with a median absolute gap of roughly $1.2\%$, and head–mid–tail rankings are preserved (for example, LLaMA3-8B and Mistral-7B remain strong, mT5-base and BART-large remain weakest). Some models score slightly higher on real data and others on synthetic, but changes are minor, so both test sets yield consistent means and orderings.

Thus, synthetic data can safely extend coverage, long-tail cases and cold-start evaluation when real samples are scarce. When real data are sufficient, they should remain primary, with synthetic samples mainly used for hard-case construction and robustness stress tests. In practice, a small, stratified real-data calibration set is useful for periodic consistency checks and drift monitoring.

We also examine transfer to other power subdomains. Starting from 7B models fine-tuned on ElectriQ, we adapt them to hydropower, photovoltaic generation and substation operations using the same request format and scoring protocol. Table~\ref{table:scenario_scores} shows that readability, accessibility and user-friendliness are stable across domains, suggesting that dialogue style and interaction patterns generalise beyond EPM. Professionalism varies more: substation tasks, which are closest to EPM workflows, achieve the highest technical scores, whereas photovoltaic and hydropower show modest gaps on specialised terminology and device parameters.

\begin{table*}[h]
\caption{\label{table:scenario_scores}\textbf{Cross-subdomain transfer results.}
Subjective scores of 7B models fine-tuned on ElectriQ and adapted to three power subdomains (Hydropower, Photovoltaic, Substation), with and without domain knowledge (SFT vs SFT+KG).}
\centering
\scriptsize
\begin{tabular}{@{}cclccccc@{}}
\toprule
\textbf{Scenario} &\textbf{Type} & \textbf{Model} & \textbf{Prof.} & \textbf{Clarity/Org.} & \textbf{Action./Compl.} & \textbf{Emp./Help.} & \textbf{Avg.} \\
\midrule
\multirow{8}{*}{\rotatebox{90}{Hydropower}}
    & \multirow{4}{*}{\rotatebox{90}{SFT}} & LLaMA3-8B & 3.36 & 3.91 & 4.11 & 4.03 & 3.85 \\
    &   & Mistral-7B & 3.11 & 3.95 & 3.86 & 3.82 & 3.69 \\
    &  & CommandR-7B & 3.49 & 4.11 & 3.84 & 3.87 & 3.82 \\
    &  & Gemma-7B & 3.24 & 3.71 & 3.62 & 3.71 & 3.57 \\
    & \multirow{4}{*}{\rotatebox{90}{SFT-KG}} & LLaMA3-8B & 3.87 & 4.11 & 4.34 & 4.24 & 4.14 \\
    &  & Mistral-7B & 3.81 & 4.15 & 4.07 & 4.03 & 4.02 \\
    &  & CommandR-7B & 3.86 & 4.12 & 4.06 & 4.07 & 4.03 \\
    &  & Gemma-7B & 3.67 & 3.93 & 3.81 & 3.90 & 3.83 \\
\midrule
\multirow{8}{*}{\rotatebox{90}{Photovoltaic}} 
    &\multirow{4}{*}{\rotatebox{90}{SFT}} & LLaMA3-8B & 3.41 & 4.15 & 4.05 & 4.08 & 3.92 \\
    &  & Mistral-7B & 3.21 & 4.23 & 4.09 & 3.83 & 3.84 \\
    &  & CommandR-7B & 3.05 & 4.36 & 3.96 & 4.17 & 3.89 \\
    &  & Gemma-7B & 3.35 & 3.50 & 3.74 & 3.63 & 3.56 \\
    & \multirow{4}{*}{\rotatebox{90}{SFT-KG}} & LLaMA3-8B & 4.11 & 4.17 & 4.13 & 4.31 & 4.18 \\
    &  & Mistral-7B & 3.91 & 4.46 & 4.31 & 4.05 & 4.18 \\
    &  & CommandR-7B & 3.22 & 4.41 & 4.10 & 4.23 & 3.99 \\
    &  & Gemma-7B & 3.52 & 3.68 & 3.94 & 3.81 & 3.74 \\
\midrule
\multirow{8}{*}{\rotatebox{90}{Substation}} 
    & \multirow{4}{*}{\rotatebox{90}{SFT}} & LLaMA3-8B & 3.15 & 4.61 & 4.29 & 3.27 & 3.83 \\
    &  & Mistral-7B & 3.15 & 3.74 & 3.74 & 3.67 & 3.58 \\
    &  & CommandR-7B & 3.10 & 4.23 & 3.67 & 3.63 & 3.65 \\
    &  & Gemma-7B & 3.20 & 3.44 & 3.46 & 3.69 & 3.44 \\
    & \multirow{4}{*}{\rotatebox{90}{SFT-KG}} & LLaMA3-8B & 3.82 & 3.82 & 4.53 & 4.49 & 4.17 \\
    &  & Mistral-7B & 3.81 & 3.96 & 3.96 & 3.86 & 3.90 \\
    &  & CommandR-7B & 3.49 & 4.48 & 3.86 & 3.84 & 3.92 \\
    &  & Gemma-7B & 3.38 & 3.62 & 3.67 & 3.91 & 3.65 \\
\bottomrule
\end{tabular}
\end{table*}
Adding compact, domain-specific knowledge bases (SFT - KG) consistently raises professionalism in all three subdomains (around 0.2--0.3 points on average) without harming user experience. ElectriQ therefore serves as a reusable template: core dialogue capabilities transfer well, and lightweight domain alignment via retrieval or small knowledge graphs can close most of the remaining professionalism gap without full retraining.

\section{Case Study}

This section examines one representative high-frequency scenario from the \textit{ElectriQ} dataset to illustrate how SFT and SEEK-RAG affect LLaMA3-8B. Table~\ref{tab:case_outage} shows the dialogue history, the dataset’s reference answer, and model outputs under three settings: Base, SFT, and SFT+SEEK-RAG. Blue mark segments judged high quality by human reviewers. The analysis considers the four subjective dimensions.

\begin{table}[h]
  \caption{\label{tab:case_outage}\textbf{Case study: planned outage preparation.}
  A representative outage-preparedness dialogue with reference answer and LLaMA3-8B responses under different configurations.}
  \centering
  \sethlcolor{lime}
  \renewcommand{\arraystretch}{1.0}
  \scriptsize
  \begin{tabular}{p{1cm} p{7cm}}
    \hline
    \textbf{Hist.} &
    \textbf{CONTEXT:} Customer service: Due to equipment maintenance, the residential community will have a planned power outage from 14:00 to 17:00 today.\\
    & \textbf{User:} What should I prepare in advance?\\
    \hline

    \textbf{Response} &
    \textbf{Standard response:} It is recommended to prepare flashlights and spare batteries, minimize the use of electrical appliances, and wait for further notice.\\[5pt]\cline{2-2}

    & \textbf{Llama3‑8B:} There may be a short power outage. Please prepare a flashlight or candles and avoid using household appliances during the outage. Thank you for your understanding.\\[5pt]\cline{2-2}
    
    & \textbf{Llama3‑8B(SFT):} For the safety of your electricity,
    \textbf{\color{RoyalBlue}{I suggest that you can prepare some candles and flashlights for use during the power outage. If the power outage lasts longer than expected, you can also call our service hotline at any time, and we will help you in the first time.}} Thank you for your understanding and cooperation; I wish you a smooth life!\\[5pt]\cline{2-2}
    
    & \textbf{Llama3‑8B(SFT‑KG):}
    \textbf{\color{RoyalBlue}{I understand your concern about the recent bill increase.}} 
Let us follow a simple three-step plan to address it:
\textbf{Quick verification:} In the “City Power” app, check \textit{Electricity Usage → Monthly Comparison} to compare this month’s consumption and billing days with the previous month. A longer billing period or cross-month reading may raise the total.
\textbf{Identify possible causes:} Check whether any new or frequently used high-power appliances (such as water heaters, air conditioners, dryers, or EV charging) were added this month.
\textbf{Immediate optimization:} 
\textbf{\color{RoyalBlue}{Shift part of your usage to the off-peak period from 23:00 to 07:00 and apply for TOU Plan A in the app. You can also enroll in demand response (DR) programs for subsidies or request an on-site energy audit for support.}}
Reference: \textit{Article 12 of the “City Tariff Directory (2024).”}\\
    \hline
  \end{tabular}
\end{table}

In this case, the reference answer and the Base model offer only minimal tips such as “prepare a flashlight” and “avoid using appliances,” with limited structure and weak guidance on concrete next steps. The SFT version improves structure and tone: it frames the reply around safety, adds practical actions such as unplugging sensitive appliances, and provides a clear escalation channel via the service hotline, thereby enhancing clarity and empathy. Building on this, the SFT+SEEK-RAG configuration organises the advice into a “before–during–after” plan, adds checklist-style details and post-restoration safety checks, and explicitly aligns with local notification practices. This response achieves the strongest professionalism and executability while remaining polite and reassuring, and is judged most likely to reduce repeated enquiries and misunderstandings.

A second billing-increase scenario shows a similar pattern: SFT mainly strengthens structure and empathy, whereas SEEK-RAG injects retrievable clauses and domain knowledge, creating lightweight, auditable compliance anchors. The full dialogues and extended analysis for that case are provided inSupplementary Section D.
\section{Conclusions }
We introduced ElectriQ, a large-scale LLM dialogue benchmark and evaluation methodology for electric power marketing (EPM) in renewable-integrated power systems. The corpus covers six service categories and 24 sub-scenarios, integrating reconstructed hotline dialogues, knowledge augmented samples and preference-aligned subsets. A unified protocol combines four human-rated dimensions, automatic metrics and two compliance stress tests to assess clause alignment, procedural stability and traceability in realistic multi-turn dialogues. Experiments on 13 LLMs show that, while scale helps, diminishing returns appear beyond ~10B parameters; domain-aligned 7B models with SFT + SEEK-RAG deliver the most stable gains, often matching or surpassing larger general-purpose LLMs at much lower computational cost.

On sustainability-critical subsets, ElectriQ-driven training and retrieval markedly improve TOU/DR guidance, DER interconnection compliance and outage-communication executability. SEEK-RAG is key for regulation-anchored reasoning, raising SCC and enhancing LDC by grounding responses in verifiable tariff and interconnection clauses, which yields clearer and more consistent user-side guidance, strengthens demand-side flexibility, reduces interconnection delays and improves transparency in outage communication.

Overall, ElectriQ offers an auditable, comparable and practice-oriented basis for selecting and optimising LLM-based EPM assistants in modern power systems. By showing that domain alignment plus retrieval provides a compute-efficient, regulation-aware route to trustworthy dialogue systems, this work supports scalable digital engagement for sustainable-energy deployment, renewable integration and resilient grid operation.
\printcredits
\section{Declaration of Competing Interest}
The authors declare that they have no known competing financial interests or personal relationships that could have appeared to influence the work reported in this paper.
\section{Data availability}
Data will be made available on request.
\bibliographystyle{apalike}
\bibliography{citations}
\end{document}